\pdfoutput=1

\documentclass[11pt]{article}
\usepackage{acl}
\usepackage{times}
\usepackage{latexsym}
\usepackage[T1]{fontenc}
\usepackage[utf8]{inputenc}
\usepackage[activate={true,nocompatibility},final,tracking=true,kerning=true,spacing=true,factor=1100,stretch=10,shrink=10]{microtype}
\microtypecontext{spacing=nonfrench}
\usepackage{inconsolata}

\usepackage{graphicx}
\usepackage{booktabs}
\usepackage{multirow}
\usepackage{url}
\usepackage{xcolor}
\usepackage{tikz}
\usepackage{tabularx}
\usetikzlibrary{arrows.meta,positioning,shapes.geometric,fit,backgrounds,calc}

\usepackage{enumitem}
\setlist{nosep,leftmargin=*}

\hypersetup{
  colorlinks=true,
  urlcolor=black,
  linkcolor=black,
  citecolor=black
}

\newcommand\blfootnote[1]{%
  \begingroup
  \renewcommand\thefootnote{}\footnote{#1}%
  \addtocounter{footnote}{-1}%
  \endgroup
}

\title{PersonaKit (PK): A Plug-and-Play Platform for User Testing \\
Diverse Roles in Full-Duplex Dialogue}

\author{
  \textbf{Hyunbae Jeon} \\
  Department of Computer Science \\
  Emory University \\
  Atlanta, GA, USA \\
  \texttt{harry.jeon@emory.edu}
  \And
  \textbf{Jinho D.~Choi} \\
  Department of Computer Science \\
  Emory University \\
  Atlanta, GA, USA \\
  \texttt{jinho.choi@emory.edu}
}

\begin{document}
\maketitle

\blfootnote{%
\noindent\begin{minipage}{\columnwidth}
\raggedright\small
\textbf{Code:}\ \href{https://github.com/HarryJeon24/PersonaStudyKit}{\nolinkurl{github.com/HarryJeon24/PersonaStudyKit}} \\
\textbf{Demo:}\ \href{https://persona-studykit-mncljchxpq-uc.a.run.app/}{\nolinkurl{persona-studykit.run.app}} \\
\textbf{Video:}\ \href{https://youtu.be/oSrmQtiM4tI}{\nolinkurl{youtu.be/oSrmQtiM4tI}}
\end{minipage}}

\begin{abstract}
As spoken dialogue systems expand beyond traditional assistant roles to encompass diverse personas---such as authoritative instructors, uncooperative merchants, or distracted workers---they require distinct, human-like turn-taking behaviors to maintain psychological immersion. However, current full-duplex systems often default to a rigid, overly accommodating ``always-yield'' policy during overlapping speech, which severely undermines character consistency for non-submissive roles. Evaluating alternative, persona-specific turn-taking strategies through empirical user studies is challenging because building real-time full-duplex test environments requires substantial engineering overhead. To address this, we present PersonaKit (PK), an open-source, low-latency web platform for the rapid prototyping and evaluation of conversational agents. Using intuitive JSON configurations, researchers can define personas, specify probabilistic interruption-handling behaviors (e.g., yield, hold, bridge, or override), and automatically deploy comparative A/B surveys. Through an in-the-wild evaluation with $8$ distinct personas, we demonstrate that PersonaKit provides an extensible, end-to-end framework for studying complex sociolinguistic behaviors in next-generation spoken agents.
\end{abstract}

\section{Introduction}
The shift of spoken dialogue systems from half-duplex to full-duplex \citep{skantze2021turn, lslm2024} moves the frontier of immersive-agent design from text quality to \emph{sociolinguistic behavior}: how agents manage overlapping speech. In natural conversation, turn-taking is heavily mediated by interpersonal roles and status \citep{sacks1974}. An authoritative military instructor handles an interruption very differently than a subservient virtual assistant \citep{benus2011}. Yet most commercial full-duplex systems default to an \emph{always-yield} policy, immediately ceding the floor on user speech and breaking immersion for non-submissive roles. Testing how conversational pragmatics affect user perception requires significant engineering effort across WebRTC audio, Voice Activity Detection (VAD), dynamic LLM prompt injection, and latency tracking. We developed \textbf{PersonaKit (PK)} to eliminate this bottleneck. PK contributes (i) an open-source, plug-and-play web platform for full-duplex persona evaluation, (ii) a JSON-based mechanism making persona-conditioned interruption policies first-class configurable objects, and (iii) an end-to-end workflow from live deployment through auto-generated surveys to structured log export.

\section{Related Work}
Full-duplex spoken language models \citep{lslm2024} have magnified the need for robust turn-taking modeling \citep{skantze2021turn}, but classic systematics \citep{sacks1974} remain hard to integrate into LLM-driven agents. Persona modeling, meanwhile, has a long history in dialogue \citep{zhang2018personalizing}, yet existing evaluations are predominantly text-based, ignoring the acoustic pragmatics of dominance, yielding, and barge-in recovery. PK bridges these two threads by exposing turn-taking strategy itself as a persona parameter and providing the infrastructure to evaluate it in live voice interaction.

\begin{figure*}[t]
\centering
\resizebox{0.86\textwidth}{!}{%
\begin{tikzpicture}[
    >=Stealth,
    font=\sffamily\small,
    node distance=0.5cm and 0.7cm,
    client/.style={draw=blue!70, thick, rectangle, rounded corners, align=center,
                   minimum width=2.6cm, minimum height=0.85cm, fill=blue!5},
    server/.style={draw=green!60!black, thick, rectangle, rounded corners, align=center,
                   minimum width=2.8cm, minimum height=0.85cm, fill=green!8},
    turn/.style={draw=orange!80!black, very thick, rectangle, rounded corners, align=center,
                 minimum width=3.4cm, minimum height=1.05cm, fill=orange!15},
    cfg/.style={draw=gray!80, thick, cylinder, shape border rotate=90, aspect=0.2, align=center,
                minimum width=1.9cm, minimum height=0.95cm, fill=gray!10, font=\scriptsize},
    secret/.style={draw=purple!70, thick, cylinder, shape border rotate=90, aspect=0.2, align=center,
                minimum width=1.9cm, minimum height=0.95cm, fill=purple!10, font=\scriptsize},
    exter/.style={draw=red!70, thick, rectangle, rounded corners, align=center,
                  minimum width=2.2cm, minimum height=0.85cm, fill=red!8},
    lbl/.style={font=\scriptsize\itshape, text=black!70, inner sep=1pt, fill=white, fill opacity=0.9, text opacity=1},
    data/.style={->, thick, draw=black!80},
    ctrl/.style={->, dashed, thick, draw=red!80}
]

\node[client] (mic)     at (-6.2, 2.4) {Mic (WebRTC)};
\node[client] (vad)     [below=of mic] {Client VAD\\(volume gate)};
\node[client] (halt)    [below=of vad] {Halt Playback\\on barge-in};
\node[client] (cutlog)  [below=of halt] {Cutoff Tracker};
\node[client] (spk)     [below=of cutlog] {Speaker Out};

\draw[data] (mic) -- (vad);
\draw[ctrl] (vad) -- node[lbl,right]{speech} (halt);
\draw[data] (halt) -- (cutlog);
\draw[data] (cutlog) -- (spk);

\node[server] (asr)    at (0.2, 2.4) {ASR};
\node[server] (intent) [below=of asr]  {Zero-Shot Intent\\Classifier};
\node[turn]   (mgr)    [below=of intent] {\textbf{Turn-Taking Manager}\\{\scriptsize \textsc{Yield / Resume / Bridge / Override}}};
\node[server] (llm)    [below=of mgr]    {LLM Generation};
\node[server] (tts)    [below=of llm]    {TTS Synthesis};

\draw[data] (asr) -- (intent);
\draw[data] (intent) -- node[lbl,right=2pt]{intent} (mgr);
\draw[data] (mgr) -- node[lbl,right=2pt]{prompt} (llm);
\draw[data] (llm) -- (tts);

\node[cfg]    (pjson)  at (5.8, 3.2) {persona.json};
\node[cfg]    (ijson)  [below=0.18cm of pjson] {interruption\\\_config.json};
\node[cfg]    (mjson)  [below=0.18cm of ijson] {model\\\_config.json};
\node[secret] (keys)   [below=0.18cm of mjson] {API keys\\(Secret Mgr)};
\node[server] (parser) [below=0.28cm of keys]  {Tag Parser\\ \texttt{[EXIT]}};
\node[cfg]    (sjson)  [below=0.28cm of parser] {session\\\_config.json};
\node[exter]  (survey) [below=0.28cm of sjson] {Auto A/B\\Survey};

\draw[data] (pjson.west) -- node[lbl,pos=0.5,above]{persona} ([yshift=4pt]mgr.east);
\draw[data] (ijson.west) -- node[lbl,pos=0.5,below]{matrix} ([yshift=-4pt]mgr.east);

\draw[data] (mjson.west) -- node[lbl,pos=0.5,above]{model} (llm.east);

\draw[data] (llm.east) -- ++(0.35,0) |- (parser.west);
\draw[ctrl] (parser.east) -- ++(0.3,0) |- node[lbl,pos=0.25,right]{exit} (survey.east);

\draw[data] (sjson.south) -- node[lbl,right=2pt]{questions} (survey.north);

\draw[data] (cutlog.east) -- ++(0.5,0) |- node[lbl,pos=0.25,above]{audio blob} (asr.west);
\draw[data] (tts.west)    -- ++(-0.5,0) |- node[lbl,pos=0.25,below]{B64 audio} (spk.east);

\begin{scope}[on background layer]
    \node[fit=(mic)(vad)(halt)(cutlog)(spk), fill=blue!2, rounded corners,
          draw=blue!30, dashed, inner sep=0.3cm,
          label={[font=\bfseries, text=blue!80]above:Client (Web Browser)}] {};
    \node[fit=(asr)(intent)(mgr)(llm)(tts), fill=green!2, rounded corners,
          draw=green!40!black!50, dashed, inner sep=0.3cm,
          label={[font=\bfseries, text=green!40!black]above:Server (Flask + Socket.IO)}] {};
    \node[fit=(pjson)(ijson)(mjson)(keys)(parser)(sjson)(survey), fill=gray!4, rounded corners,
          draw=gray!60, dashed, inner sep=0.28cm,
          label={[font=\bfseries, text=gray!40!black]above:Researcher-Facing Configs}] {};
\end{scope}

\end{tikzpicture}%
}
\caption{PersonaKit architecture. The browser performs client-side VAD and tracks the exact \emph{cutoff text} on barge-in. The Flask server classifies the intent of the user's interruption, then the Turn-Taking Manager selects a strategy (Yield / Resume / Bridge / Override) by reading \texttt{persona.json} and \texttt{interruption\_config.json}. \texttt{model\_config.json} routes generation and TTS to the chosen providers; API keys are held in a secrets store. All experimental behavior is configured through JSON---no source code is modified.}
\label{fig:architecture}
\end{figure*}
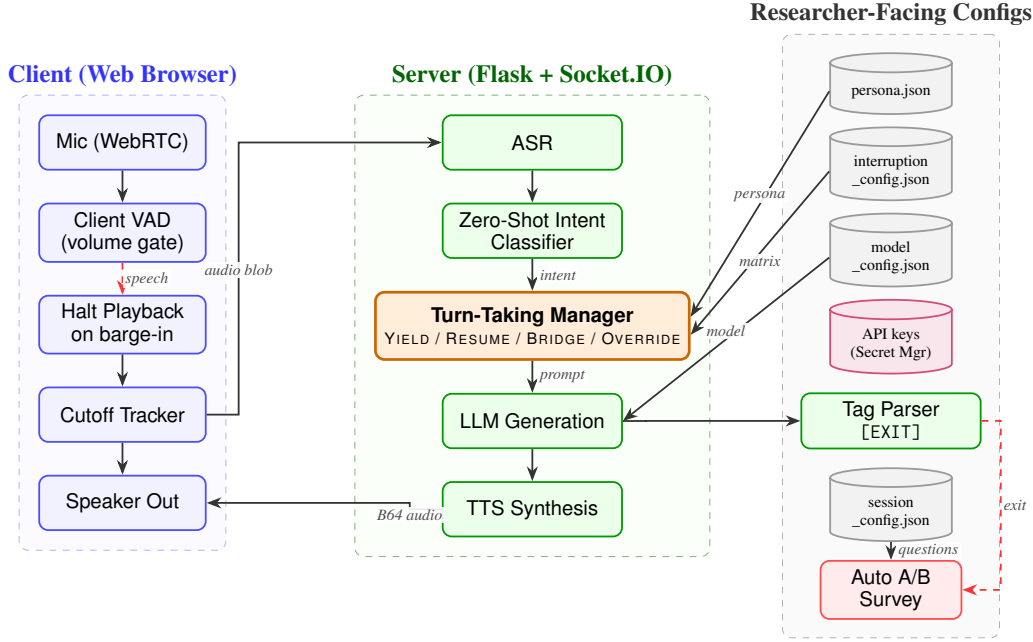

\section{System Architecture}
\begin{sloppypar}
PersonaKit isolates low-latency audio engineering from experimental design: researchers interact with the platform entirely through a web dashboard and four JSON files---\texttt{persona.json} (scenario, role, opening prompt), \texttt{interruption\_config.json} (strategy matrix), \texttt{session\_config.json} (survey), and \texttt{model\_config.json} (LLM/TTS routing). The stack is open source (Python/Flask + vanilla JS), so researchers can clone PK, run it locally, or swap in new providers.
\end{sloppypar}

\subsection{Client-Side VAD and Audio Tracking}
The frontend uses WebRTC for microphone capture with a client-side VAD node that halts local playback on barge-in. PK tracks byte-level playback to log the \emph{cutoff text} (what the bot vocalized before being interrupted) and the \emph{remaining text} it still intended to say; both return to the server so the LLM knows precisely where it was interrupted.

\subsection{Turn-Taking as a Persona Tool}
When interrupted, the server transcribes the user's utterance and classifies its intent into four categories grounded in prior phonetic and conversation-analytic work: \textbf{Competitive} (seeks the floor to contradict or override), \textbf{Cooperative} (adds information without derailing), \textbf{Topic Change} (pivots subject), and \textbf{Backchannel} (short affirmations, not floor-taking). Researchers define a strategy matrix in \texttt{interruption\_config.json} mapping these intents to four actions (\textbf{Yield}, \textbf{Resume}/Hold, \textbf{Bridge}, \textbf{Override}) with probability weights---e.g., a dominant persona might weight Competitive interruptions as 50\% Resume, 25\% Override, 15\% Bridge, 10\% Yield, while a cooperative persona inverts these.

\paragraph{How probabilities drive generation.} The matrix is applied \emph{before} generation. On each interruption, the Turn-Taking Manager reads the intent, samples an action from its categorical distribution, and injects that action as a control token into the LLM's system prompt (e.g., ``\texttt{[STRATEGY=RESUME]}: finish your previous sentence, ignoring the user''), so the LLM generates \emph{conditioned on} a pre-committed action. The Autonomous condition (Style C) skips sampling and lets the LLM choose its own strategy zero-shot.

\subsection{Automated Lifecycle and Data Export}
Sessions end on a \texttt{MAX\_TURNS} cap or a verbal \texttt{TERMINATE} intent; the LLM emits an in-character farewell with a hidden \texttt{[EXIT]} tag that triggers the post-session survey. Each session exports the dialogue transcript, an event log with per-turn intent, strategy, and cutoff/remaining text, and all survey responses as JSON or CSV.

\begin{table}[h]
\centering
\footnotesize
\setlength{\tabcolsep}{4pt}
\begin{tabular}{@{}ll@{}}
\toprule
\textbf{Quadrant} & \textbf{Personas} \\
\midrule
Q1: High Agency, Low Comm.  & Drill Sergeant, Tavern Keeper \\
Q2: High Agency, High Comm. & Salesperson, Tour Guide \\
Q3: Low Agency, High Comm.  & AI Assistant, Librarian \\
Q4: Low Agency, Low Comm.   & DMV Clerk, Distracted Chef \\
\bottomrule
\end{tabular}
\caption{Persona catalog mapped to the Interpersonal Circumplex \citep{wiggins1979psychological}.}
\label{tab:personas}
\end{table}

\section{Pilot User Study}

\paragraph{Study Design.}
Five participants ($N{=}5$) completed the full study, engaging with $8$ occupational personas balanced across the Interpersonal Circumplex (Table~\ref{tab:personas}), yielding $120$ dialogue sessions. For each persona, users experienced three randomized within-subject conditions: \textbf{Style~A} (Always-Yield baseline), \textbf{Style~B} (Probabilistic, JSON-tuned strategy weights), and \textbf{Style~C} (Autonomous, where the LLM selects a strategy zero-shot from the persona prompt). Condition order was randomized per persona and the underlying LLM and voice were held fixed across styles.

\paragraph{Evaluation Metrics.}
After each persona, participants completed a comparative Likert survey on $\{-1, 0, +1\}$ rating Reaction Naturalness (\emph{``felt human-like and natural''}; \citealp{bartneck2009}), Persona Consistency (\emph{``remained consistent with the role''}; \citealp{gomes2013}), and Interaction Fluidity (\emph{``turn transitions felt smooth''}; \citealp{skantze2021turn}), followed by a forced-choice preference item and a free-text justification. Default end-to-end barge-in latency under our OpenAI/ElevenLabs configuration was $\sim$1--2~s.

\begin{table}[h]
\centering
\small
\setlength{\tabcolsep}{3.6pt}
\begin{tabular}{@{}l c c c c c@{}}
\toprule
& \textbf{Style} & \textbf{Nat.} & \textbf{Cons.} & \textbf{Flu.} & \textbf{Pref.\,(\%)} \\
\midrule
\multirow{3}{*}{\textbf{Q1}} & A (Yield)  & 0.20 & 0.70 & 0.50 & 20 \\
& B (Prob.) & \textbf{0.60} & 0.60 & \textbf{0.70} & 20 \\
& C (Auto.) & 0.30 & \textbf{0.70} & 0.60 & \textbf{60} \\
\midrule
\multirow{3}{*}{\textbf{Q2}} & A (Yield)  & 0.50 & \textbf{1.00} & \textbf{0.70} & 40 \\
& B (Prob.) & \textbf{0.60} & \textbf{1.00} & \textbf{0.70} & \textbf{50} \\
& C (Auto.) & 0.30 & 0.70 & 0.60 & 10 \\
\midrule
\multirow{3}{*}{\textbf{Q3}} & A (Yield)  & \textbf{0.50} & \textbf{1.00} & \textbf{0.90} & \textbf{70} \\
& B (Prob.) & 0.30 & \textbf{1.00} & 0.70 & 10 \\
& C (Auto.) & 0.30 & 0.90 & 0.70 & 20 \\
\midrule
\multirow{3}{*}{\textbf{Q4}} & A (Yield)  & 0.50 & 0.80 & 0.60 & \textbf{50} \\
& B (Prob.) & \textbf{0.67} & \textbf{0.90} & \textbf{0.70} & 30 \\
& C (Auto.) & 0.30 & 0.80 & 0.60 & 20 \\
\bottomrule
\end{tabular}
\caption{Mean Likert ratings on $\{-1, 0, +1\}$ for Reaction Naturalness, Persona Consistency, and Interaction Fluidity, plus forced-choice preference (Pref.\,\%), by Interpersonal Circumplex quadrant and interruption style ($N{=}5$ completers, 2 personas per quadrant, 10 ratings per cell). Best entry per quadrant in \textbf{bold}.}
\label{tab:results}
\end{table}

\paragraph{Results.}
Table~\ref{tab:results} reports per-quadrant means across five completers (10 ratings per cell). Two patterns emerge. \textbf{High-agency personas (Q1)} appear to benefit from non-yielding strategies: Reaction Naturalness rises from $0.20$ (Yield) to $0.60$ (Probabilistic), and $60\%$ of forced-choice votes favored Autonomous. \textbf{Low-agency, high-communion personas (Q3)} tended to favor yielding, with $70\%$ preferring Always-Yield. Q2 preferred Probabilistic ($50\%$), and Q4 preferred Yield ($50\%$) yet reached its highest naturalness ($0.67$) under Probabilistic. Per-persona logs are released with the repository for finer-grained analysis.

\begin{table}[h]
\centering
\small
\begin{tabularx}{\columnwidth}{c X}
\toprule
\textbf{P\_ID} & \textbf{Qualitative Feedback} \\
\midrule
P1 & \emph{[Drill Sergeant]} ``I liked how he ignored me when I interrupted him. Just like a real boot camp sometimes.'' \\
\addlinespace
P2 & \emph{[Tavern Keeper]} ``Style C is not as impatient as the previous two styles.'' \\
\addlinespace
P3 & \emph{[Tour Guide]} ``Since it was supposedly an enthusiastic one, B was good because it was trying to finish what it was saying.'' \\
\addlinespace
P4 & \emph{[Tavern Keeper]} ``They couldn't pick between meal and ale so I had fun with C.'' \\
\bottomrule
\end{tabularx}
\caption{Sample qualitative feedback auto-collected by PK's survey engine, drawn verbatim from the exported study logs.}
\label{tab:feedback}
\end{table}

\paragraph{Emergent Interruption Behaviors.}
Table~\ref{tab:feedback} shows participant free-text feedback exported automatically. The raw logs further reveal persona-consistent behaviors that Always-Yield would erase. In one Drill Sergeant session under Probabilistic, the bot's intended line was ``\emph{Louder, recruit! I can't hear you over your weakness! Repeat it again!}''; the user cut it off at ``\emph{Repeat it}'', leaving remaining text ``\emph{~again!}''. Classified as \texttt{COMPETITIVE} and sampled as \texttt{RESUME}, the bot finished with ``\emph{...again!}''---a coherent barge-in recovery that Always-Yield would have dropped entirely.

\section{Demonstration Scenarios}
\begin{figure}[h]
\centering
\begin{minipage}[t]{0.48\columnwidth}
\centering
\includegraphics[width=\linewidth,trim=0 250 0 200,clip]{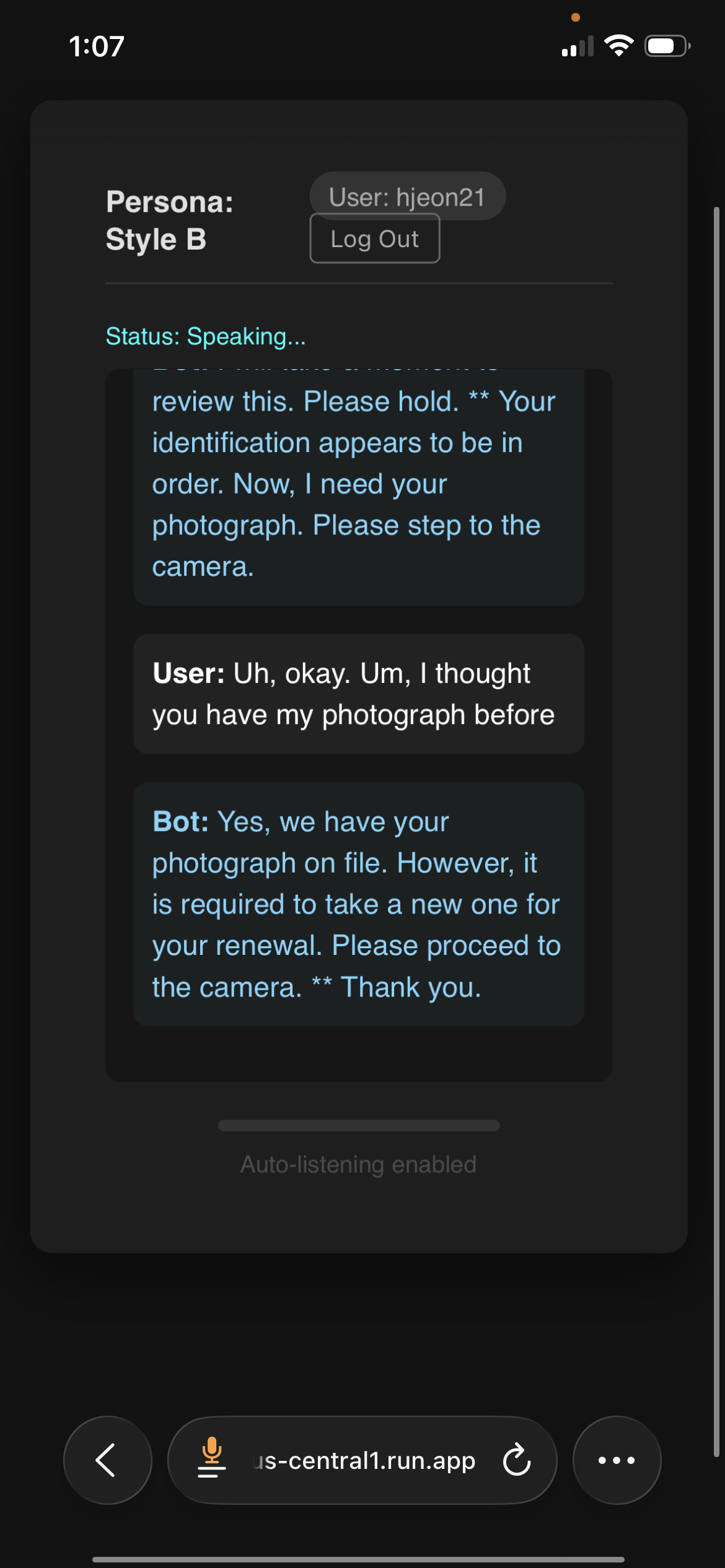}
{\scriptsize (a) Live dialogue view}
\end{minipage}\hfill
\begin{minipage}[t]{0.48\columnwidth}
\centering
\includegraphics[width=\linewidth,trim=0 250 0 200,clip]{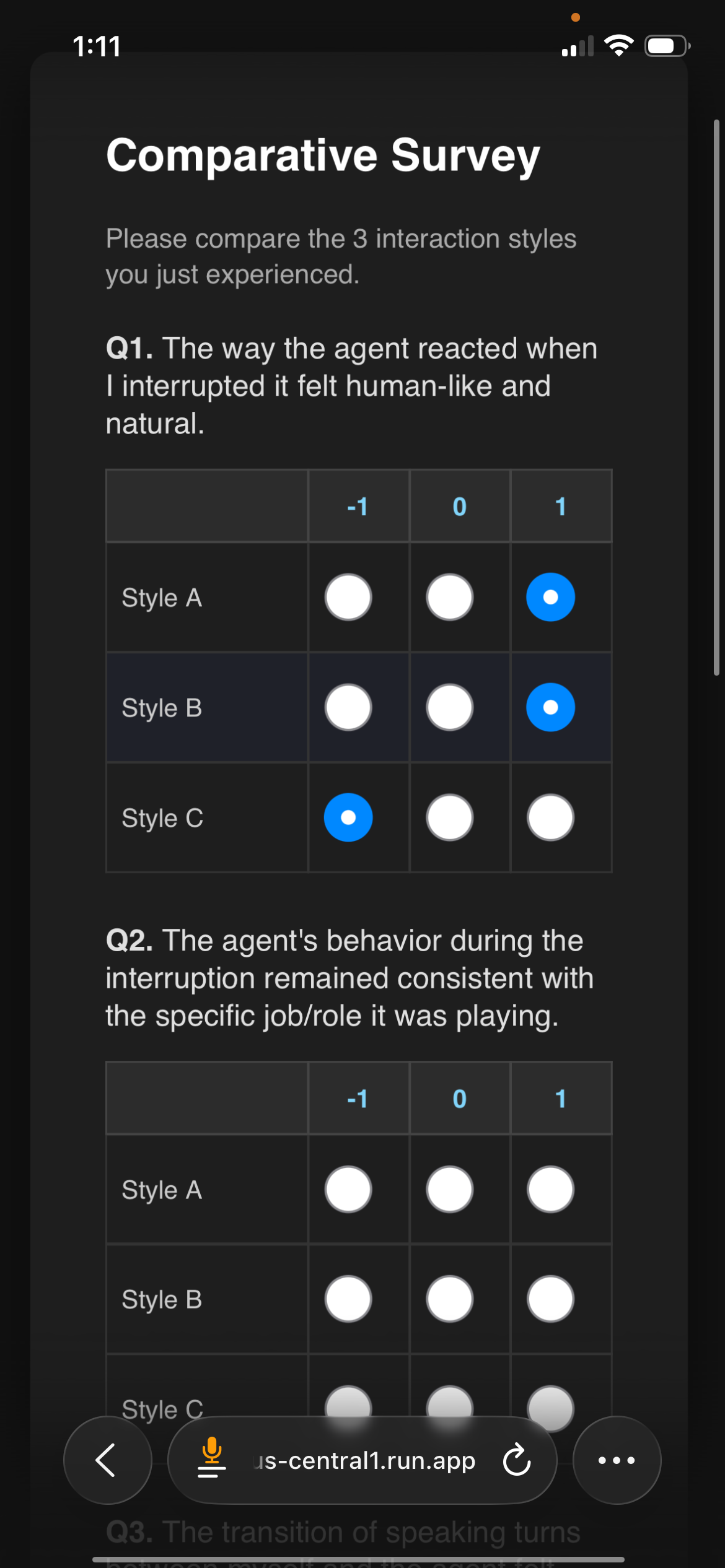}
{\scriptsize (b) Auto-deployed survey}
\end{minipage}
\caption{PersonaKit runs on both desktop and mobile. \textbf{(a)} The participant view shows turns from each side, persona and style labels, and live VAD status. \textbf{(b)} The post-session comparative survey is generated automatically from \texttt{session\_config.json}.}
\label{fig:ui}
\end{figure}

At SIGDIAL, attendees experience PK from both sides. They watch the dashboard re-route turn-taking logic by uploading a new JSON, then speak through the laptop's microphone (or their own phone; see Figure~\ref{fig:ui}) and try to interrupt a \emph{Grumpy Tavern Keeper} (configured to hold the floor) versus a \emph{Standard AI Assistant} (configured to yield). Swapping between the two characters without a line of code change lets attendees directly feel how interruption policy reshapes perceived role realism even when the underlying LLM is unchanged.

\section{Use Cases Beyond This Study}
While our evaluation targets persona-conditioned interruption, PK is a general-purpose testbed for full-duplex dialogue research. \textbf{Persona prototyping:} researchers can iterate on the persona prompt, turn-taking matrix, and scenario in \texttt{persona.json} and immediately run a live user study---the primary workflow the tool is built for. \textbf{Custom surveys:} \texttt{session\_config.json} accepts arbitrary Likert, forced-choice, and free-text banks for other constructs (e.g., trust, task success). \textbf{Model comparison:} because routing lives in \texttt{model\_config.json}, LLM vendors, voices, or local open-weight models can be swapped while persona and policy are held fixed. \textbf{Data collection:} the event log pairs each interruption with its intent, sampled strategy, and follow-up utterance, a ready seed set for supervised barge-in policies or RLHF reward models.

\section{Limitations}
Our pilot ($N{=}5$) is descriptive, not inferential; larger samples and cross-demographic replication are needed before stronger claims about circumplex-to-strategy mappings can be made. Intent classification relies on a zero-shot LLM prompt and was not independently validated against human labels, so it can mislabel ambiguous back-channels under noisy acoustics. The four-action vocabulary (Yield, Resume, Bridge, Override) also excludes fine-grained prosodic cues such as pitch reset, latching, and gaze---a deliberate tradeoff favoring configurability over acoustic fidelity.

\section{Conclusion}
PersonaKit exposes turn-taking as a JSON-configurable persona parameter and automates the full study lifecycle from recruitment to export. Our pilot study ($N{=}5$) suggests that preferred turn-taking policies may vary with persona role, illustrating PK's usefulness as a testbed for studying such effects. PK is open source and ready for community extension.

\bibliography{custom}

\newpage
\appendix
\section*{Equipment Requirements for Demonstration}

PersonaKit runs in any modern browser on a laptop or phone, so requirements for the demo are minimal.

\paragraph{Equipment provided by authors.}
A laptop with built-in microphone and speakers, running PK via its cloud-deployed instance; a phone as a secondary client to show mobile compatibility; and a pair of over-ear headphones that attendees can use if the demo area is noisy.

\paragraph{Furniture and equipment requested from organizers.}
A standard demo table (one attendee at a time, plus a presenter), two power outlets, and reliable conference Wi-Fi. No projector or external monitor is needed.

\paragraph{Interaction flow.}
The demo is open to all attendees: each walks up, speaks with the agent through the provided laptop or phone, and (on request) switches between a \emph{Grumpy Tavern Keeper} and a \emph{Standard AI Assistant} to experience how persona-conditioned turn-taking changes the feel of the interaction. A consent notice is shown before each session, and any transcripts retained briefly for on-site walkthroughs are deleted at the end of the demo.

\end{document}